\def\year{2021}\relax
\documentclass[letterpaper]{article}
\usepackage{aaai21}
\usepackage{times}
\usepackage{helvet}
\usepackage{courier}
\usepackage{graphicx}
\usepackage{natbib}  

\title{Towards Trustworthy Cross-patient Model Development}
\author {Ali El-Merhi\textsuperscript{\rm 1,2},  
Helena Odenstedt Herg{\'e}s\textsuperscript{1,2},
        Linda Block\textsuperscript{\rm 1,2}, 
        Mikael Elam\textsuperscript{\rm 3}, 
        Richard Vithal\textsuperscript{\rm 1,2}, 
        Jaquette Liljencrantz\textsuperscript{\rm 3},
        Miroslaw Staron\textsuperscript{\rm 4} \\
}
\affiliations{
    \textsuperscript{\rm 1}Sahlgrenska Academy, Institute of Clinical Sciences, University of Gothenburg, Gothenburg, Sweden, \\ 
    \textsuperscript{\rm 2} Department of anesthesia and intensive care, Sahlgrenska university hospital, Gothenburg, Sweden, \\
    \textsuperscript{\rm 3} Institute of Neuroscience and physiology, Sahlgrenska Academy, University of Gothenburg, \\
    \textsuperscript{\rm 4} Computer Science and Engineering, Chalmers $|$ University of Gothenburg, Gothenburg, Sweden
}

\begin{document}
\maketitle


\begin{abstract}
\begin{quote}
Machine learning is used in medicine to support physicians in examination, diagnosis, and predicting outcomes. One of the most dynamic area is the usage of patient generated health data from intensive care units. The goal of this paper is to demonstrate how we advance cross-patient ML model development by combining the patient's demographics data with their physiological 
data. We used a population of patients undergoing Carotid Enderarterectomy (CEA), where we studied differences in model performance and explainability when trained for all patients and one patient at a time. The results show that patients' demographics has a large impact on the performance and explainability and thus trustworthiness. We conclude that we can increase trust in ML models in a cross-patient context, by careful selection of models and patients based on their demographics and the surgical procedure. 
\end{quote}
\end{abstract}

\section{Introduction}
\label{sec:introduction}
\noindent
Critical care units increasingly often adopt machine learning to improve the outcome of patients \cite{maslove2021}, \cite{flechet2016informatics}. Critical care patients and patients undergoing surgery are connected to equipment used by physicians to monitor, such signals as ECG, blood pressure or blood oxygenation. These signals can also be used to predict upcoming adverse events and to decrease the time needed to detect these events, thereby reducing the number of invasive tests or increasing the certainty of the diagnosis \cite{wijnberge2020effect}, \cite{carra2020data}. 

One of the challenges when using ML models in critical care is how to use a model trained on one patient for diagnosing another patient \cite{komorowski2019artificial}. From the medical perspective, the demographics of the patients can differ greatly, even if they undergo the same procedure or suffer from the same disease. From the technical perspective, the signals collected during the procedures can be collected differently (e.g. variations in how electrodes are attached to the patient). Therefore, in order to trust the ML models, it is important to identify the variability, understand the source of it and then choose the right strategy to handle it.

In practice, in most cases, ML engineers and scientists adopt the strategy to collect more data in the hope to create models which are robust to these variations \cite{smiti2020machine}. This means that the models' predictions will be of a better quality, but, at the same time, more data can lead to lower explainability and thus trustworthiness. 

In this paper, we set off to investigate how much the results vary from when we train the model on all patients, and when we train the model on one patient at a time. The goal was to explore the variability of models and therefore assess how robust the models are to different patients. 

The data was collected from a pipeline where we designed and conducted a clinical study of patients undergoing CEA \cite{staron2021robust}, \cite{block2020cerebral}.  
We used a RandomForest classifier, which has been used in many studies analyzing signals from patients \cite{qi2012random}.  

Our results show that there is a large variability between how we train and use the model. For the same model, when trained on the data from all patients, the accuracy was 0.86. However, for the same algorithm, but trained on one patient at a time, the accuracy varied from 0.07 to 0.63. The model performed best for the pair of patients who were the most similar (same gender, similar age, similar initial status). Although the results were better for the model trained on all patients, we could observe that the model is more trustworthy if we can trace the training model to a specific patient rather than a large dataset with high variability. 

\section{Context -- CEA and Collected Signals}
\label{sec:context}
\noindent
Carotid endarterectomy (CEA) is recommended in symptomatic patients with $\geq$ 50\% stenosis and in asymptomatic patients with $\geq$ 60\% stenosis \cite{eckstein2018european}. CEA poses an increased risk of cerebral ischemia due to embolization and hemodynamic derangement during clamping of the carotid arteries. In our centre, all carotid endarterectomies are performed with patients under general anaesthesia. All intraoperative monitoring, including invasive arterial blood pressure (ABP) and neuromonitoring, is connected to patients and monitoring starts prior to induction of anaesthesia.  Baseline values before induction of anesthesia are recorded. Noradrenaline infusion is titrated as needed to maintain ABP close to baseline values. After the skin incision and surgical exposure but before plaque removal, the surgeon cross-clamps the common carotid artery , the external carotid artery , and the internal carotid artery (ICA), so that the stenotic bifurcation is excluded from the circulation.  At this moment, the surgeon, taking into account information from available neuromonitoring and measured blood pressure in the distal stump of the ICA, evaluates the risk of cerebral ischemia due to hypoperfusion and decides if a shunt is to be placed to maintain cerebral perfusion during clamping. After plaque removal the artery is sutured and the clamp is released to restore blood flow. After emergence from anaesthesia, the patient undergoes frequent bedside neurological examinations. 

Since the patients are connected to monitors, we can collect the following signals, and transform them into ML features:
\begin{itemize}
    \item EEG: frequency bands (alpha, beta, delta, gamma, and theta) for each channel (F3-Cz, F4-Cz, C3-Cz, C4-Cz, P3-Cz, and P4-Cz). 
    \item ECG: Inter-beat intervals (IBI), BPM (Heartbeats per minute), SDNN (Standard deviation of the NN (R-R) intervals), SDSD (standard deviation of successive differences), RMSSD (Root mean square of the successive differences), PNN50 (proportion of NN50\footnote{The number of pairs of successive NN (R-R) intervals that differ by more than 50 ms.} divided by the total number of NN (R-R) intervals), PNN20 (The proportion of NN50\footnote{The number of pairs of successive NN (R-R) intervals that differ by more than 20 ms.} divided by the total number of NN (R-R) intervals), HR\_MAD (median absolute deviation of RR intervals), SD1 (standard deviation perpendicular to identity line), SD2 (standard deviation along identity line), S (Area of ellipse described by SD1 and SD2), SD1/SD2 (SD1/SD2 ratio), Breathing rate (estimated from the ECG signal), 
    \item ABP (Arterial Blood Pressure): ABP mean, 
    \item NIRS (Near-Infrared Spectroscopy): rSO2 left (side of the brain), and rSO2 right (side of the brain), and 
    \item SpO\textsubscript{2} (Oxygen Saturation): SpO\textsubscript{2} value. 
\end{itemize}

The data set contains also a decision class, which is the set of events which we want to recognize/diagnose during CEA. We use the following events:
\begin{itemize}
  \item Pre-clamp anaesthesia -- induction of anaesthesia- when the patient’s anesthesia starts. 
  \item Pre-clamp surgery -- opening the cartoid artery. 
 \item Clamped artery -- when the carotid artery is closed.
 \item Shunt -- when the surgeons make a shunt around the closed artery to restore blood flow (in case the patient's blood oxygenation decreases to a pre-defined level).
  \item HRV window anasthesia pre-clamp -- six breaths during one minute in the ventilator, to establish a baseline for the analysis of HRV signals under controlled conditions. This is done both before the clamping of the artery and after the clamping of the artery. 
  \item Post-operative anesthesia -- when the patient is regaining consciousness.
  \item Post-operative care -- when the patient is awake and under post-operative care in the neurointensive care unit. 
\end{itemize}

The dataset in our study consists of 48 feature columns and one class column. It 

We experiment with several ML algorithms -- RandomForest, SVM, AdaBoost, Decision Trees (CART) and artifical neural networks. We chose RandomForest for our comparisons due to its robustness and ability to handle large data sets \cite{qi2012random}. In this study, we follow the procedure:
\begin{enumerate}
    \item Collect the data from all patients in one table.
    \item Train and validate the classifier based on all data, using 0.33 train-test split. 
    \item Train the classifier on one patient at a time and validate on all other patients. 
\end{enumerate}

This procedure let us compare how the performance of the classifier varies depending on the way in which it is trained and evaluated.

\section{Results and interpretation}
\label{sec:results}
\begin{figure}[!htb]
    \centering
    \includegraphics[width=\columnwidth]{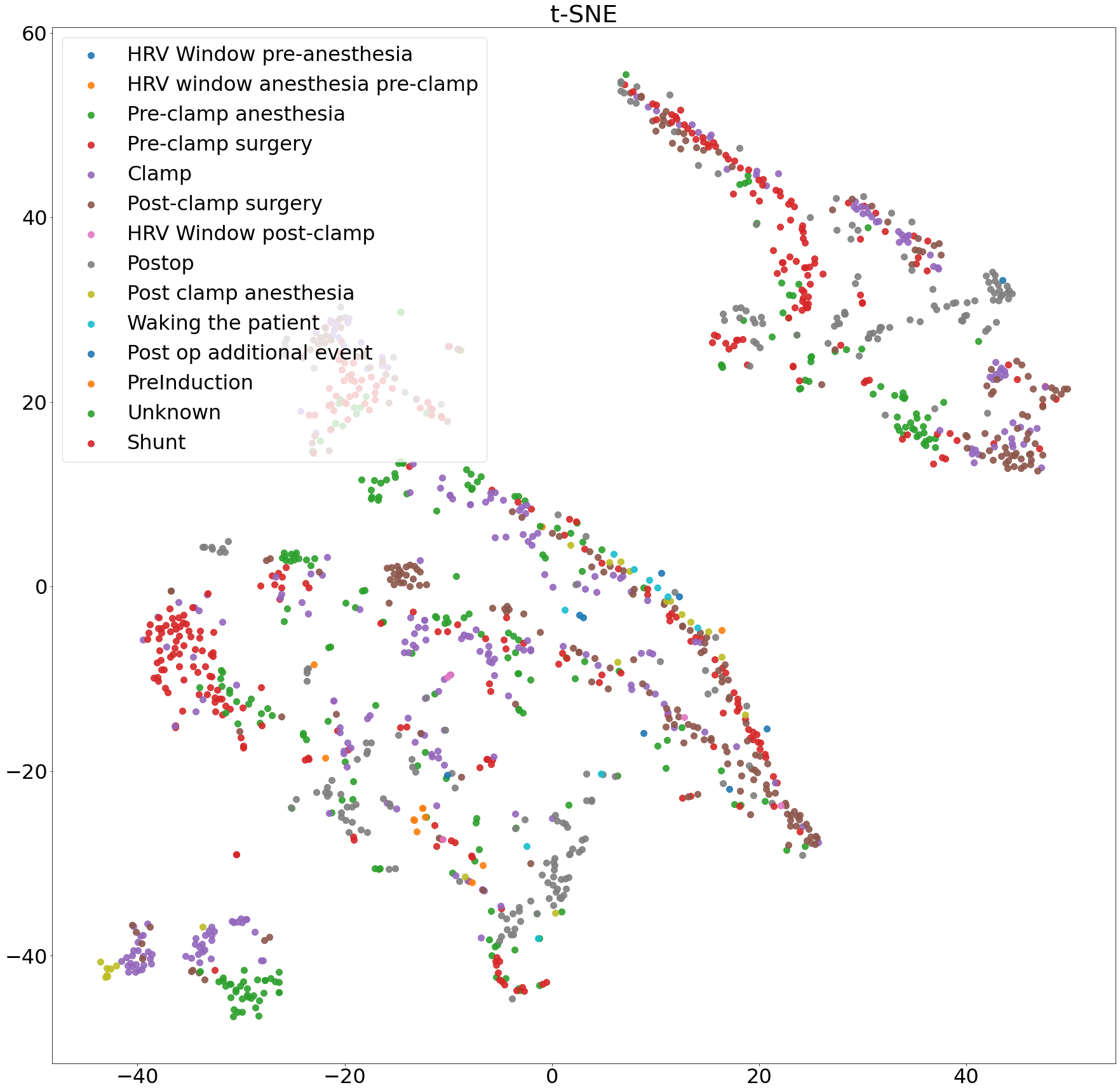}
    \caption{t-SNE diagram to illustrate the data set. Each data point represents one minute of a CEA procedure. Each color represents the type of event/decision class.}
    \label{fig:tsne}
\end{figure}

\noindent
To understand the data, we use the t-SNE diagram as a means for visualization of all data points (Figure \ref{fig:tsne}). The diagram shows that there are areas where different events/decision classes overlap, but in the majority of cases the data points for specific events are grouped together (e.g. the purple dots, corresponding to the clamping of the artery, forms several distinct areas in the figure). 

\begin{figure}[!htb]
    \centering
    \includegraphics[width=\columnwidth]{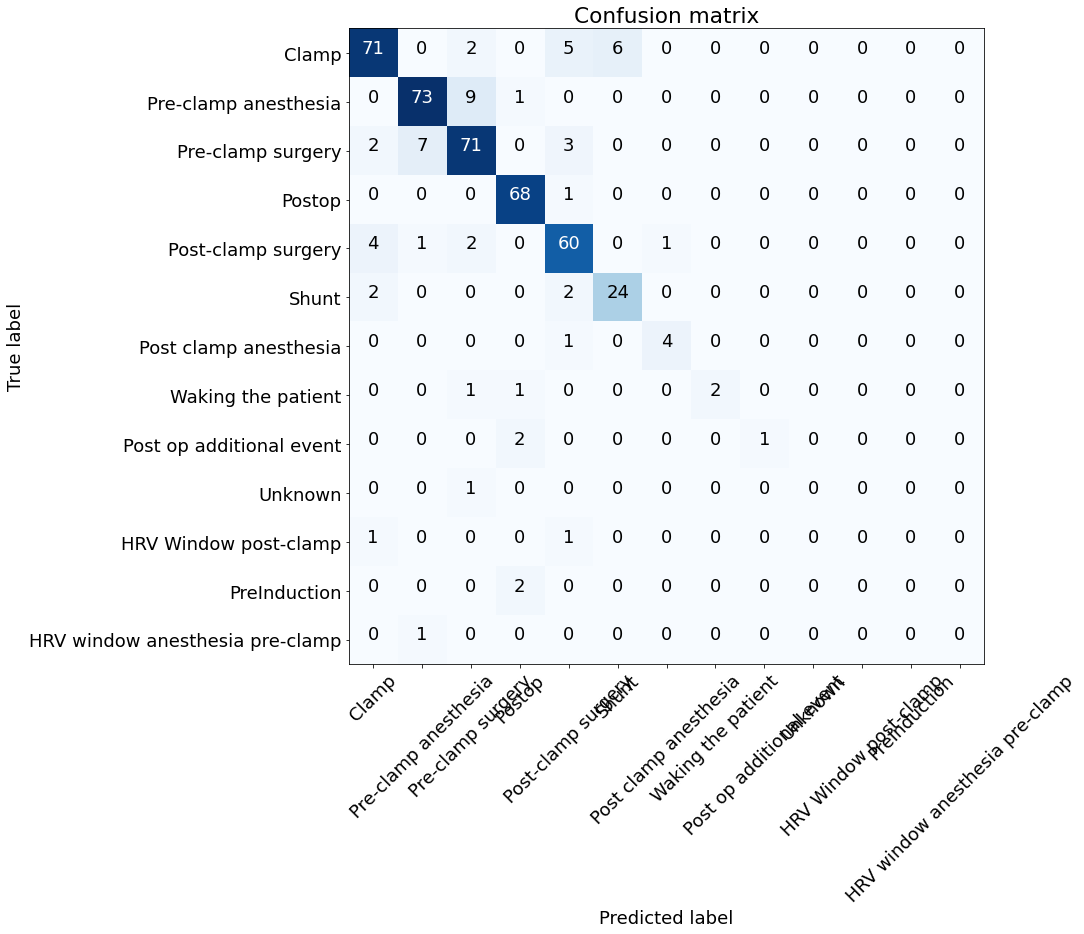}
    \caption{Confusion matrix for the cross-patient model for the validation data (0.33 of the entire data set).}
    \label{fig:confusion}
\end{figure}

We train the classifier with the following parameters: number of trees: 64, number of leaves: 128. The accuracy for that model is 0.86. Figure \ref{fig:confusion} presents the confusion matrix for this model. The confusion matrix shows that most of the errors are for these classes which have the fewest data points (e.g. Pre-induction); this is despite balancing classes using weights. 

\begin{figure}[!htb]
    \centering
    \includegraphics[width=\columnwidth]{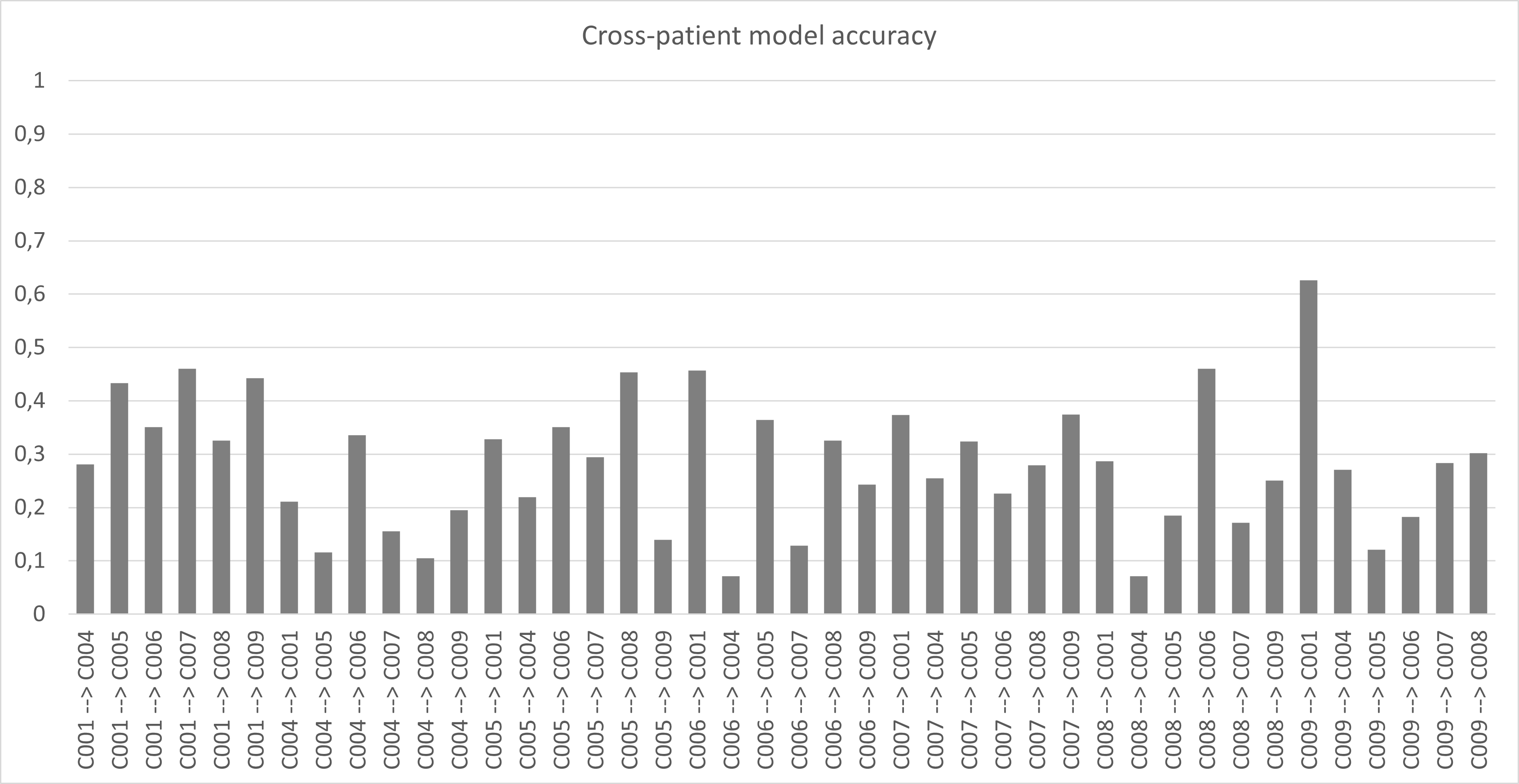}
    \caption{Accuracy of the models trained for each pair of patients. The model is trained on one patient and then validated on another patient (train patient --\textgreater validation patient).}
    \label{fig:cross_patient}
\end{figure}

Figure \ref{fig:cross_patient} presents a diagram of accuracy for a cross-patient trained model. Each model has been trained on one patient and applied on another. The lowest accuracy is 0.07 and the highest is 0.63. Although the diagram shows a significant variability in accuracy, we can understand it better if we observe the difference between the feature importance for the cross-patient model and the model for the best pair of patients (C009 and C001).  

\begin{figure}[!htb]
    \centering
    \includegraphics[width=\columnwidth]{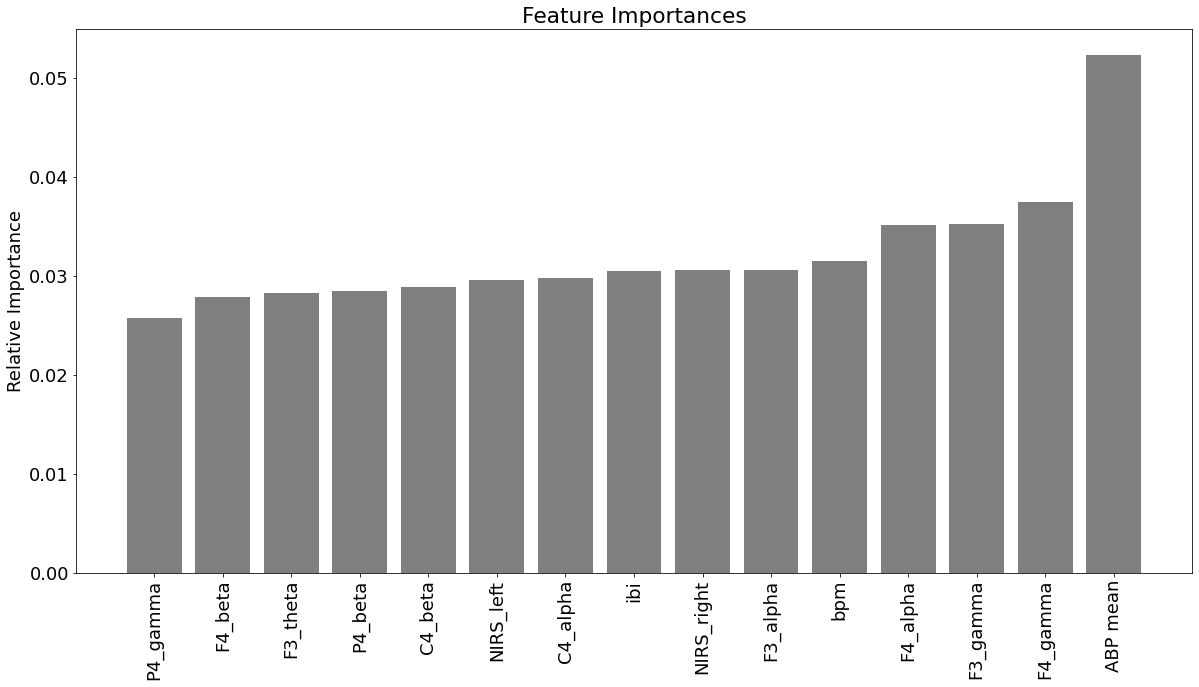}
    \caption{Feature importance for the model trained for all patients. The most important feature is the Arterial Blood Pressure (ABP).}
    \label{fig:all_patients}
\end{figure}

From the perspective of explainability and, thus, trust in the model, accuracy is not the best validation metric. In our study, we use feature importance charts as a means to understand how good the models are and whether they can be trusted. For the cross-patient model, Figure \ref{fig:all_patients} presents the feature importance chart for the most important 15 features. The chart shows that it is the arterial blood pressure (ABP) which is the most important feature. 

\begin{figure}[!htb]
    \centering
    \includegraphics[width=\columnwidth]{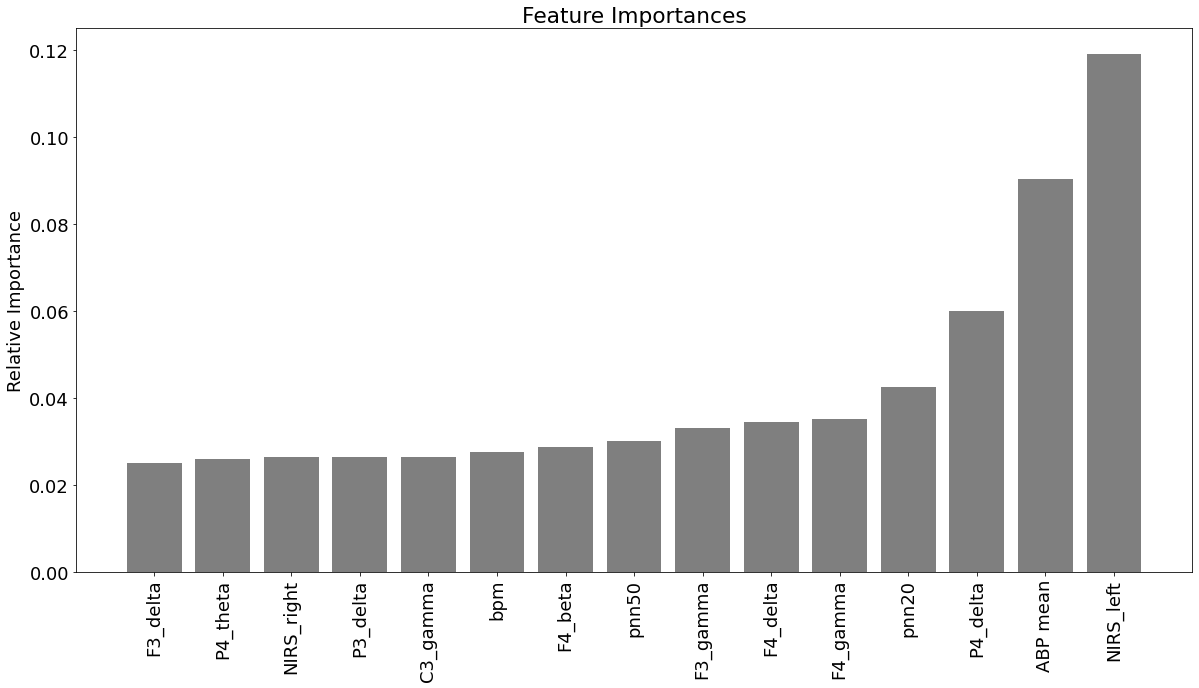}
    \caption{Feature importance for the model trained for patient C009 and applied on the patient C001 (the pair with the highest accuracy or 0.63). The most important feature is the NIRS signal.}
    \label{fig:best_patients}
\end{figure}

For the model trained on individual patients, however, the feature importance chart is different -- see Figure \ref{fig:best_patients}. The most important feature is the NIRS signal from the left side of the head. Although both ABP and NIRS can have a good explanation, there is a medical reason for this difference. Patients C001 and C009, are similar in terms of: which artery was clamped (right), condition for admittance (Amaurois fugax dx; Paresthesia left hand vs. Hemiparesis (left); Amaurois fugax sin.), gender and age. Patients C006 and C004 (the worst match) are of different gender and arterial clamp was performed on different sides. As NIRS has been found useful for same-side arterial clamp \cite{lewis2018cerebral}, it confuses the algorithm for different-side arterial clamping, thus rendering this signal untrustworthy in this context. 

Considering both of these charts together, our interpretation is that the similarity of the patients condition is important for the trustworthiness of the model -- clamping on the same side strengths the models accuracy, compare to other pairs of patients. We observe also that the number of data points in the data set increases the accuracy of the model in general.

\begin{figure}[!htb]
    \centering
    \includegraphics[width=\columnwidth]{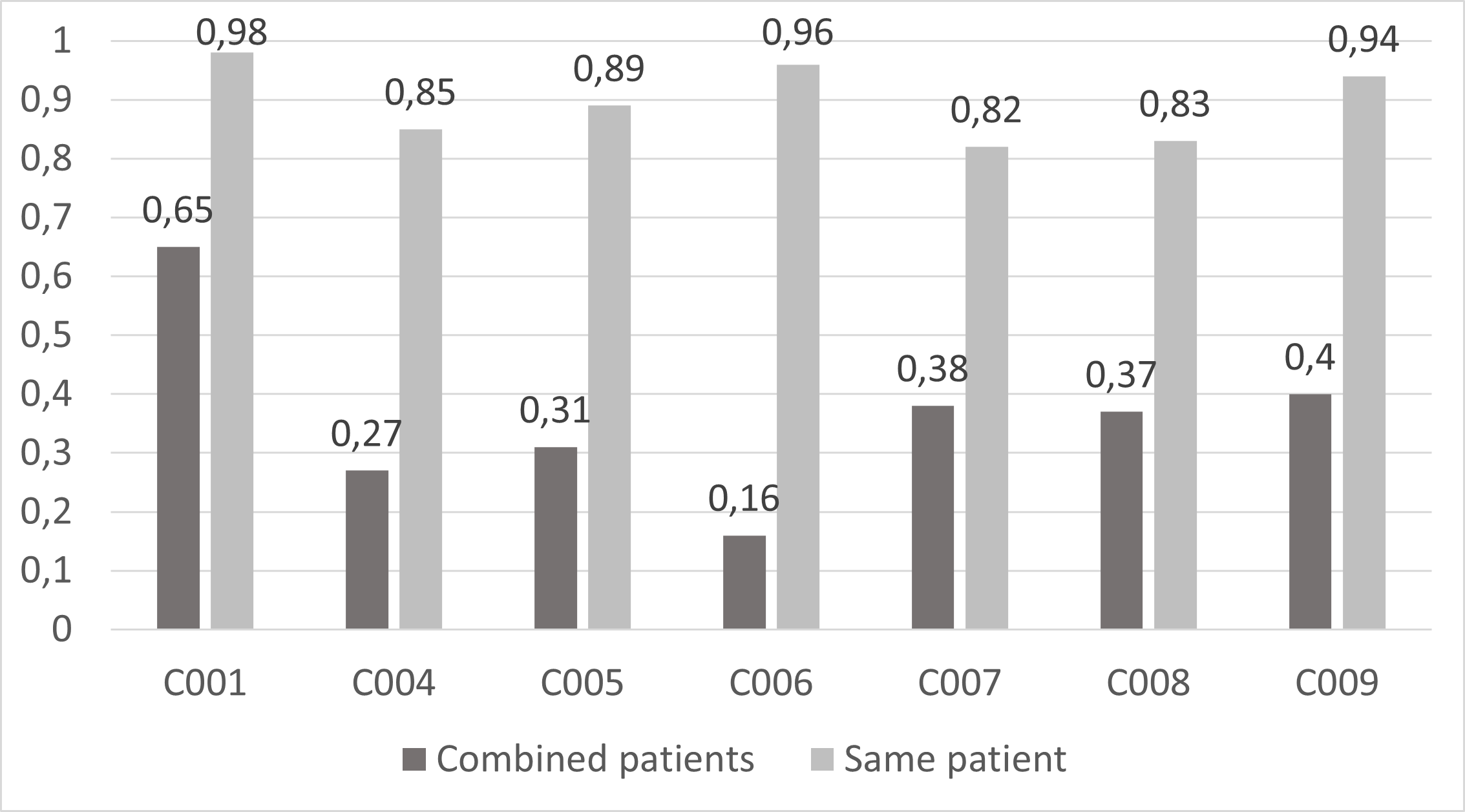}
    \caption{Accuracy of the combined model -- one patient is set aside for validation while the model is trained for the data from all other patients; and models trained on the same patient -- setting aside 0.33 points for validation.}
    \label{fig:combined}
\end{figure}

We combine these two approaches: set aside one patient for validation while training the model on all other patients. The results are shown in Figure \ref{fig:combined}. 

\begin{figure}[!htb]
    \centering
    \includegraphics[width=\columnwidth]{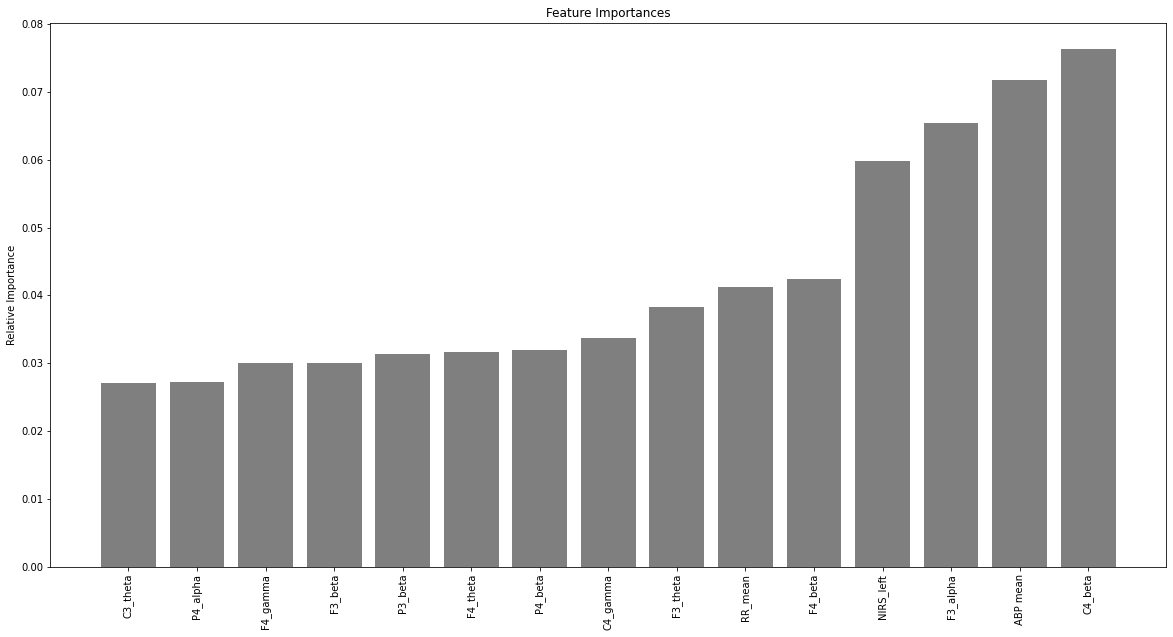}
    \caption{Feature importance chart for the model validated for patient C001.}
    \label{fig:features_combined}
\end{figure}

The highest accuracy is for patient C001 -- 0.65. The lowest accuracy is for patient C006 -- 0.16. There is an improvement across all patients, although not a large one. For patient C001, we also include the feature importance chart -- Figure \ref{fig:features_combined}. 

Our recommendation, therefore, is to decide early whether we want more explainability or higher accuracy. For a reference, the accuracy if we train the model for 0.67 of the data for one patient and validate on the remaining 0.33, the accuracy is above 0.82.  

\section{Conclusions}
\label{sec:conclusions}
\noindent
In the context of using machine learning models in medical practice, trustworthiness is one of the crucial components in the adoption of this technology in practice. In this paper, we present results of a study on how the performance of a Random Forest classifier changes when using the same data in two different ways -- training the model on the entire data set and training it per patient. 

The results show that the model trained on all data is more accurate, but has lower explainability and thus trustworthiness. The model trained on a single patient has lower accuracy, but higher explainability. The accuracy depends on the selection of the patients, which is, therefore, a required feature for this algorithm. 

Our conclusions are that we need to combine both approaches -- selecting the most relevant patients and training the model on a larger data set. We plan to evaluate these conclusions based on new patients groups. 

\bibliography{aaai}

\begin{thebibliography}{11}
\providecommand{\natexlab}[1]{#1}
\providecommand{\url}[1]{\texttt{#1}}
\providecommand{\urlprefix}{URL }
\expandafter\ifx\csname urlstyle\endcsname\relax
  \providecommand{\doi}[1]{doi:\discretionary{}{}{}#1}\else
  \providecommand{\doi}{doi:\discretionary{}{}{}\begingroup
  \urlstyle{rm}\Url}\fi

\bibitem[{Block et~al.(2020)Block, El-Merhi, Liljencrantz, Naredi, Staron, and
  Odenstedt~Herg{\`e}s}]{block2020cerebral}
Block, L.; El-Merhi, A.; Liljencrantz, J.; Naredi, S.; Staron, M.; and
  Odenstedt~Herg{\`e}s, H. 2020.
\newblock Cerebral ischemia detection using artificial intelligence (CIDAI)—A
  study protocol.
\newblock \emph{Acta Anaesthesiologica Scandinavica} 64(9): 1335--1342.

\bibitem[{Carra et~al.(2020)Carra, Salluh, da~Silva~Ramos, and
  Meyfroidt}]{carra2020data}
Carra, G.; Salluh, J.~I.; da~Silva~Ramos, F.~J.; and Meyfroidt, G. 2020.
\newblock Data-driven ICU management: Using big data and algorithms to improve
  outcomes.
\newblock \emph{Journal of Critical Care} .

\bibitem[{Eckstein(2018)}]{eckstein2018european}
Eckstein, H.-H. 2018.
\newblock European Society for Vascular Surgery guidelines on the management of
  atherosclerotic carotid and vertebral artery disease.
\newblock \emph{European Journal of Vascular and Endovascular Surgery} 55(1):
  1--2.

\bibitem[{Flechet, Grandas, and Meyfroidt(2016)}]{flechet2016informatics}
Flechet, M.; Grandas, F.~G.; and Meyfroidt, G. 2016.
\newblock Informatics in neurocritical care: new ideas for Big Data.
\newblock \emph{Current Opinion in Critical Care} 22(2): 87--93.

\bibitem[{Komorowski(2019)}]{komorowski2019artificial}
Komorowski, M. 2019.
\newblock Artificial intelligence in intensive care: are we there yet?
\newblock \emph{Intensive care medicine} 45(9): 1298--1300.

\bibitem[{Lewis et~al.(2018)Lewis, Parulkar, Bebawy, Sherwani, and
  Hogue}]{lewis2018cerebral}
Lewis, C.; Parulkar, S.~D.; Bebawy, J.; Sherwani, S.; and Hogue, C.~W. 2018.
\newblock Cerebral neuromonitoring during cardiac surgery: a critical appraisal
  with an emphasis on near-infrared spectroscopy.
\newblock \emph{Journal of cardiothoracic and vascular anesthesia} 32(5):
  2313--2322.

\bibitem[{Maslove, Elbers, and Clermont(2021)}]{maslove2021}
Maslove, M.~D.; Elbers, P.~W.; and Clermont, G. 2021.
\newblock Artificial intelligence in telemetry: what clinicians should know.
\newblock \emph{Intensive Care Medicine} (3): 143--151.

\bibitem[{Qi(2012)}]{qi2012random}
Qi, Y. 2012.
\newblock Random forest for bioinformatics.
\newblock In \emph{Ensemble machine learning}, 307--323. Springer.

\bibitem[{Smiti(2020)}]{smiti2020machine}
Smiti, A. 2020.
\newblock When machine learning meets medical world: Current status and future
  challenges.
\newblock \emph{Computer Science Review} 37: 100280.

\bibitem[{Staron et~al.(2021)Staron, Herg{\'e}s, Naredi, Block, El-Merhi,
  Vithal, and Elam}]{staron2021robust}
Staron, M.; Herg{\'e}s, H.~O.; Naredi, S.; Block, L.; El-Merhi, A.; Vithal, R.;
  and Elam, M. 2021.
\newblock Robust Machine Learning in Critical Care--Software Engineering and
  Medical Perspectives.
\newblock \emph{arXiv preprint arXiv:2103.08291} .

\bibitem[{Wijnberge et~al.(2020)Wijnberge, Geerts, Hol, Lemmers, Mulder, Berge,
  Schenk, Terwindt, Hollmann, Vlaar et~al.}]{wijnberge2020effect}
Wijnberge, M.; Geerts, B.~F.; Hol, L.; Lemmers, N.; Mulder, M.~P.; Berge, P.;
  Schenk, J.; Terwindt, L.~E.; Hollmann, M.~W.; Vlaar, A.~P.; et~al. 2020.
\newblock Effect of a machine learning--derived early warning system for
  intraoperative hypotension vs standard care on depth and duration of
  intraoperative hypotension during elective noncardiac surgery: the HYPE
  randomized clinical trial.
\newblock \emph{Jama} 323(11): 1052--1060.

\end{thebibliography}

\end{document}